# Enhancing Canine Musculoskeletal Diagnoses: Leveraging Synthetic Image Data for Pre-Training AI-Models on Visual Documentations

**Martin Thißen[1], Thi Ngoc Diep Tran[1], Ben Joel Schönbein[1], Ute Trapp[1], Barbara Esteve Ratsch[2], Beate Egner[2], Romana Piat[1], and Elke Hergenröther[1]**

[1]Darmstadt University of Applied Sciences, Darmstadt, 64295, Germany
[2]Veterinary Academy of Higher Learning, Babenhausen, 64832, Germany

**ABSTRACT**

The examination of the musculoskeletal system in dogs is a challenging task in veterinary practice. The careful diagnosis as well as the evaluation of very complex findings is getting increasingly important. Therefore, a novel method has been developed that enables efficient documentation of a dog's condition through a visual representation. However, since the visual documentation is new, there is no existing training data. The objective of this work is therefore to mitigate the impact of data scarcity in order to develop an AI-based diagnostic support system that can provide veterinarians with accurate predictions. To this end, the potential of synthetic data that mimics realistic visual documentations of diseases for pre-training AI models is investigated. Specifically, this work explores whether pre-training an AI model with synthetic data can improve the overall accuracy of canine musculoskeletal diagnoses.

We propose a method for generating synthetic image data that mimics realistic visual documentations. Initially, a basic dataset containing three distinct classes is generated, followed by the creation of a more sophisticated dataset containing 36 different classes. Both datasets are used for the pre-training of an AI model, adapting it to the domain of visual documentations. Subsequently, an evaluation dataset is created, consisting of 250 manually created visual documentations for five different diseases. This dataset, along with a subset containing 25 examples, serves as the basis for evaluating the efficacy of pre-training an AI model on synthetic data.

The obtained results on the evaluation dataset containing 25 examples demonstrate a significant enhancement of approximately 10% in diagnosis accuracy when utilizing generated synthetic images that mimic real-world visual documentations. However, these results do not hold true for the larger evaluation dataset containing 250 examples, indicating that the advantages of using synthetic data for pre-training an AI model emerge primarily when dealing with few examples of visual documentations for a given disease. This implies that the use of synthetic data may not be necessary for diseases with many visual documentation examples.

Overall, this work provides valuable insights into mitigating the limitations imposed by limited training data through the strategic use of generated synthetic data, presenting an approach applicable beyond the canine musculoskeletal assessment domain.

**INTRODUCTION**

In the ever-evolving landscape of artificial intelligence (AI) applications, effectively utilizing AI models in practical use cases where training data is limited remains a challenging problem. Particularly, in veterinary medicine where the availability of huge annotated datasets is limited (Appleby and Basran 2022). One promising way to address this challenge is the exploration of few-shot learning (FSL) models, which aim to efficiently learn representations when provided only with a handful of training examples. However, state-of-the-art FSL models almost always aim to transfer knowledge gained from large-scale source tasks, utilizing huge datasets, to smaller-scale target tasks with limited training data availability (Hu et al. 2022). This approach, however, becomes problematic when the target task (e.g., canine musculoskeletal diagnosis) significantly differs from the available source tasks with huge datasets.

Therefore, an alternative strategy involves augmenting training data with synthetic examples mimicking the limited available data (Trabucco et al. 2023, Burg et al. 2023). This way, Deep Learning-based classification models can potentially thoroughly learn the characteristics of canine musculoskeletal diagnoses. For this reason, we investigated in this work whether pre-training an AI model with synthetic data can improve the overall accuracy of canine musculoskeletal diagnoses.

To this end, we developed a method to generate synthetic images mimicking visual documentations of canine musculoskeletal diagnoses. For this, we first generated a basic dataset containing three distinct classes, followed by creating a more sophisticated dataset with 36 classes. Both datasets were used to pre-train a deep learning classification model, adapting it to the domain of canine musculoskeletal diagnosis. To assess the pre-training effectiveness, we manually created an evaluation dataset containing 250 visual documentations of five diagnoses (50 examples per diagnosis). Additionally, we explored the effectiveness with minimal data, using a subset of 25 examples from the evaluation dataset (5 examples per diagnosis).

Our results indicate that the use of synthetic data is not universally beneficial. Although a significant enhancement of approximately 10% in diagnosis accuracy was achieved in a few-shot scenario (5 examples per diagnosis), this improvement did not hold true for the larger evaluation dataset with 250 examples (50 examples per diagnosis). This indicates that the advantages of using synthetic data for pre-training an AI model emerge primarily when dealing with a few examples of visual documentations for a given disease.

**GROWING DEMAND FOR EFFECTIVE DOCUMENTATION IN VETERINARY MEDICINE**

In response to the growing demand for evidence-based veterinary care, central veterinary hospitals are emerging as hubs where specialized practitioners collaborate across disciplines. This trend mirrors developments in human medicine, enabling the integration of specialties such as Physical and Rehabilitative Medicine (PRM) into veterinary practice (European Physical and Rehabilitation Medicine Bodies Alliance 2018). As the importance of PRM in veterinary medicine rises, addressing specific indications becomes imperative.

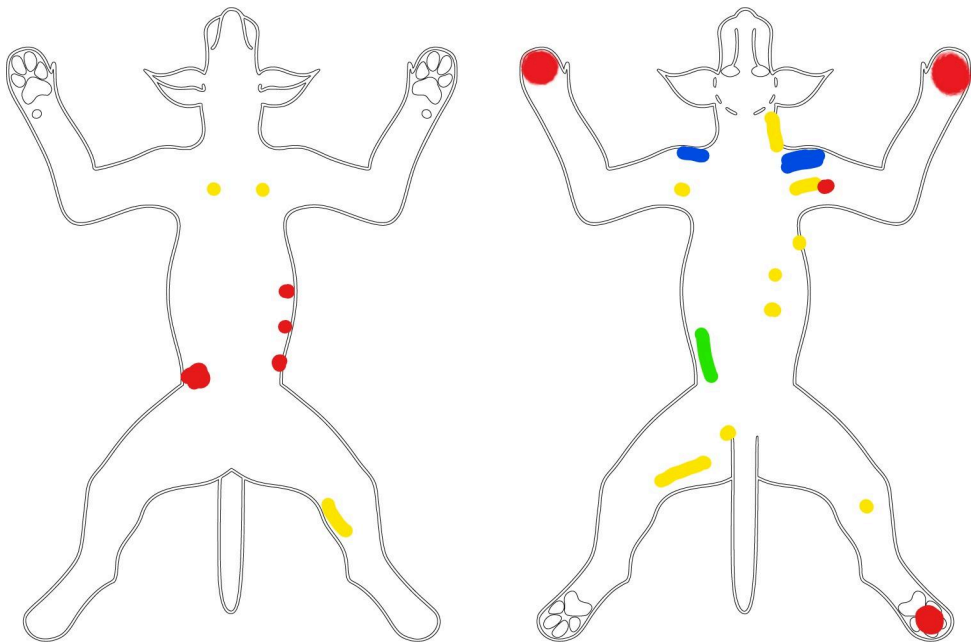

**Figure 1:** Example image of a filled-out body map (BnE © 2023 VBS GmbH) provided by Dr. Esteve Ratsch.

The escalating prevalence of sportive dogs requiring preventive and demanding soft tissue injury treatments underscores the need for enhanced PRM competence. Additionally, the aging pet population, expected to reach 30-40%, requests a proactive senior medicine approach (Fortney 2012). The rising concern of obesity, a prevalent medical issue in dogs, further emphasizes the urgency for comprehensive and effective veterinary interventions (Ratsch et al. 2022).

PRM, a discipline focused on preventing, diagnosing, treating, and rehabilitating patients with disabilities, plays a vital role in understanding the interdependencies between bodily functions and structures. The PRM practitioner collaborates with various specialists to deliver accurate diagnoses, requiring strong communication skills and interdisciplinary cooperation (European Physical and Rehabilitation Medicine Bodies Alliance 2018).

Diagnosing PRM-related issues involves detailed anamnesis, physical examinations, and functional assessments, encompassing pain, joint mobility, muscle power, and more. In complex cases, additional circulatory function testing may be necessary. A thorough diagnosis forms the foundation for a multimodal treatment plan, emphasizing the need for close interdisciplinary collaboration and effective communication (European Physical and Rehabilitation Medicine Bodies Alliance 2018).

Current veterinary tools lack efficiency in documenting and analyzing complex clinical information, hindering effective interdisciplinary communication. Body maps© (see Figure 1), commonly used in dental and dermatological practices, offer a visual solution to represent complex clinical findings (Cornwall 1992, Tarr et al. 2011, Galloway 2011). Integrating Artificial Intelligence (AI) into body maps© can expedite diagnosis and treatment planning, providing valuable insights in a time-efficient manner.

Summarizing, the integration of PRM into veterinary medicine is crucial for addressing the evolving healthcare needs of companion animals. Body maps©, supported by AI, offer a powerful tool for visualizing and communicating complex clinical findings. This innovative approach enhances interdisciplinary collaboration and ensures accurate diagnosis and treatment planning.

## BODY MAP© FOR EFFICIENT DOCUMENTATION

In order to facilitate the utilization of body maps© as an effective documentation tool for veterinarians, we have created a specialized application. By having many veterinarians use our application in the future, we aim to accumulate a substantial volume of data (body maps©). This data can then be utilized to enhance classification models, ultimately contributing to improved support for diagnosis and treatment. However, at the moment there are only a few body maps© filled out by veterinarians available.

Most practices use a practice management system (PMS) for their records (Gysin et al. 2019). Hence, all information relevant to AI models should be imported from individual practice management systems. Practice management systems are usually run on desktop computers. Therefore, our application should run on a desktop system as well. However, it's hard to draw or document inside a body map© using a desktop computer (or more specifically using a mouse). Thus, a cross-platform framework is needed. Currently, we use Flutter (Flutter - Build apps for any screen 2024) as it supports desktop, Windows, and mobile devices. In addition, our application is offline-ready, as the veterinarian might examine dogs at their home. Figure 2 shows the general architecture of our system: The backend of our application (one instance in each practice) imports relevant data from the PMS. The veterinary analyses all data related to the patient and then uses a tablet to sketch the current examination findings. All AI models run on a cloud server.

Currently, we are not able to import data from practice management systems, because these are closed systems. However, we are optimistic to find solutions and collaborate with established vendors in the future. Also, the application is still in the alpha stage and intensive UX evaluation with different veterinarians is needed. However, it is possible to document findings in the body map© and to compare body maps© using the application.

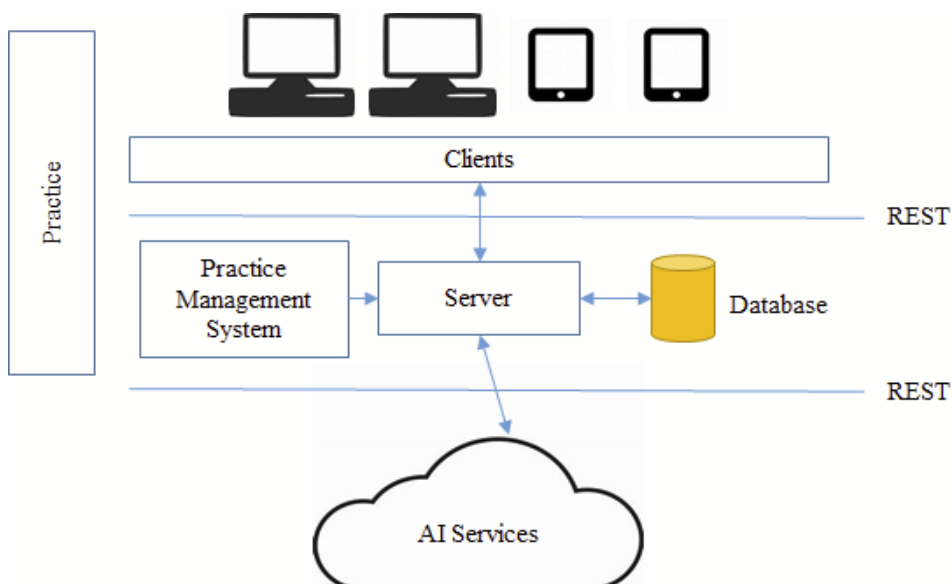

**Figure 2:** The architecture of our specialized application, allowing veterinarians to easily fill out body maps© while being compatible with existing PMSes.

## GENERATING SYNTHETIC DATA

Currently, only a handful of body maps© filled out by veterinarians (see Figure 1) are available. This is primarily due to the novelty of using a body map© for documenting canine musculoskeletal diagnoses, coupled with the fact that our developed application is currently in the alpha stage. For this reason, we developed a method for generating synthetic image data that mimics realistic visual documentations.

First, we developed an algorithm with which we can generate any number of synthetic images. Here, images of three different classes were generated, which are line, dashed line, and point cluster. These three different types will be used in the future to indicate the severity of a clinical finding made by a veterinarian. Our motivation in generating these three types was that a model trained on these synthetic images would already be able to discriminate between the three types of lines and thus acquire prior knowledge. The different classes are drawn randomly inside the body map© shown in Figure 1.

For the line and dashed line, first, a starting point *s* is randomly selected within the body map©. Then an end point *e* within the radius of 200 pixels is also randomly selected within the body map©. The lines are then generated either as a quadratic or cubic Bézier curve, whereby the control points have a random deviation in the radius of 30 pixels compared to the corresponding points on the linear line between starting point *s* and endpoint *e*.

For the point cluster, the number of points $n$ is first selected at random in the range from 3 to 20. Then a starting point $s_0$ is randomly selected and drawn inside the body map©. After that, the next point is randomly selected within a radius of 20 pixels around the previous point, which is $s_0$ in the first round. This procedure is performed $n-1$ times to generate a point cluster containing $n$ points inside the body map©.

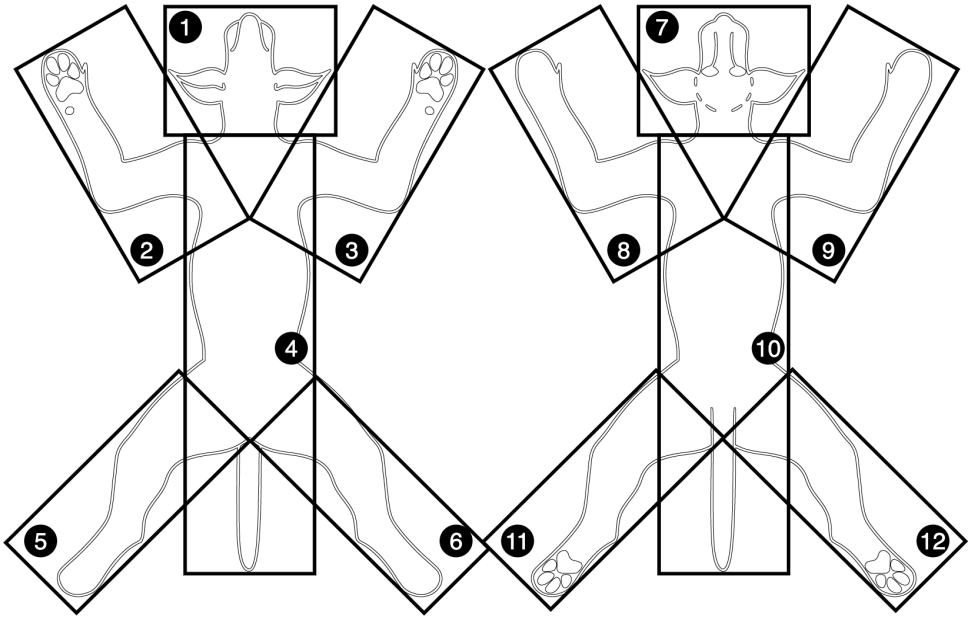

**Figure 3:** Visualization of the body map (BnE © 2023 VBS GmbH) divided into twelve regions.

We then extended our algorithm by dividing the body map© into 12 regions, as shown in Figure 3. Here we generated the three classes using the same approach as before, but this time only within one of the twelve regions. In this way, we obtained 36 different classes. The motivation for this approach was that a model trained on these synthetic images would be able to develop spatial awareness of the body map©.

Then we manually created an evaluation dataset to assess the effectiveness of using the generated synthetic data to improve the overall accuracy of canine musculoskeletal diagnoses. To this end, we defined five specific diagnoses and specified their characteristics. The five diagnoses are pelvic contusion, atrophy/hypertrophy at the forelimb, atrophy/hypertrophy at the hindlimb, low blood pressure, and high blood pressure. The characteristic of pelvic contusion, for example, is pain (indicated by the color red) in the pelvic area. For each of the five diagnoses, we manually filled out 50 body maps© for training and 10 body maps© for testing according to the specified characteristics. In addition, we extracted a subset of 25 training (5 body maps© per class) and 20 test examples (4 body maps© per class) to evaluate the effectiveness of synthetic data in a scenario where very few examples are available.

After this, we needed to determine which classification model should be used to determine the effectiveness of using synthetic data to improve the overall accuracy of canine musculoskeletal diagnoses. To this end, we examined several classification models in an exploratory analysis. After an exploratory analysis, we decided to proceed with the EfficientNet V2-S model (Tan and Le 2021) in subsequent experiments.

We then fine-tuned the EfficientNet V2-S model (pre-trained on the ImageNet dataset (Deng et al. 2009)) on the two generated datasets, the one with three classes and the one with 36 classes. In this way, we obtained two fine-tuned EfficientNet V2-S models, which we abbreviate as V2-S-3 and V2-S-36. To evaluate the effectiveness of the synthetic data, we further fine-tuned these models on the training data of our evaluation dataset. Here we have fine-tuned the V2-S-3, the V2-S-36, and the original V2-S model. We also fine-tuned the V2-S model to be able to compare it with the V2-S-3 and V2-S-36 models. This allows us to assess the effectiveness of pre-training an AI model with synthetic data on the overall accuracy of canine musculoskeletal diagnoses.

## RESULTS

To assess the effectiveness of using synthetic data, we compared two models trained with synthetic data to a model trained without synthetic data. More specifically, we fine-tuned all three models V2-S, V2-S-3, and V2-S-36 (introduced in the section before) using an evaluation dataset, which we manually created. We also extracted a subset of the evaluation dataset to evaluate the effectiveness of synthetic data in a scenario where very few examples are available.

**Table 1:** Comparison between one model trained without synthetic data (V2-S) and two models trained with synthetic data (V2-S-3 and V2-S-36). The comparison of the smaller evaluation dataset containing 25 training images is shown in the left table. The right table shows comparison results on the larger evaluation dataset containing 250 training images.

| Model | Accuracy (%) | Model | Accuracy (%) |
|---|---|---|---|
| V2-S | 80 | **V2-S** | **96** |
| **V2-S-3** | **90** | V2-S-3 | 94 |
| V2-S-36 | 80 | **V2-S-36** | **96** |
| a) | | b) | |

For the smaller evaluation dataset containing 25 training and 20 test images, the V2-S-3 model achieves an accuracy of 90%, which is 10% better than the accuracy achieved by the V2-S as well as the V2-S-36 model. The results are shown in Table 1a). These results can be interpreted such that using synthetic data leads to an overall higher accuracy of canine musculoskeletal diagnoses. However, this interpretation is only true for the V2-S-3 but not for the V2-S-36 model, showing that the use of synthetic data does not necessarily lead to a higher accuracy. Nevertheless, the improved accuracy for the V2-S-3 model by using synthetic data is promising for a scenario where only a few filled-out body maps© are available, which is currently the case in our project.

Table 1b) shows the results for the larger evaluation dataset containing 250 training and 50 test images. Here, contrastingly the V2-S and V2-S-36 models achieve a better accuracy with 96% compared to the V2-S-3 model with 94%. This contrast in achieved accuracy for the smaller and larger evaluation datasets needs to be further investigated in future work. Overall, it can be seen that more training images lead to higher accuracy. Also, the observation in Table 1a) that using synthetic images can lead to an overall improved accuracy of canine musculoskeletal diagnoses does not hold true for the larger evaluation dataset, as can be seen in Table 1b).

**DISCUSSION**

In this work, we investigated whether pre-training an AI model with synthetic data can improve the overall accuracy of canine musculoskeletal diagnoses. Our results indicate that the use of synthetic data is beneficial in scenarios where only limited data is available, specifically five examples per class in our case. For an evaluation dataset with 50 training images per class, the use of synthetic data did not lead to an improvement in the accuracy of canine musculoskeletal diagnoses.

Using an algorithmic approach to generate synthetic training data did allow us only to generate images that can represent simple features of a filled-out body map©. However, more powerful deep learning based models like generative adversarial networks (Goodfellow et al. 2014) or diffusion models (Ho et al. 2020) could potentially allow the generation of more sophisticated images that mimic body maps© filled out by veterinarians in a closer way. Nevertheless, our results show that even simplistic synthetic images can lead to improved accuracy. Furthermore, we believe that the integration of existing scientific literature in the form of text data as prior knowledge about canine musculoskeletal diagnoses is promising.

**RELATED WORK**

According to Gysin et al., most veterinarians in Switzerland use practice management systems (PMS) to administer their animal records. However, most PMSes are closed systems, and individual contracts are needed to get access to the patient health data inside these systems. Also, there is no standard way for documenting veterinary medical records, which easily causes a vendor lock-in for veterinarians. More specifically, there is hardly any structured data, most information is stored as free text (Lustgarten et al. 2020). Nevertheless, there are publications that extract information from free text and try to classify it (Sheng et al. 2022). To overcome the lack of standardization, using body maps© as a unified way of documenting a canine's musculoskeletal condition is promising.

When faced with limited available training data, there are usually two ways to mitigate the impact on an AI model's classification results. One way is to improve the AI model algorithmically, altering its architecture so that the model is able to learn generalizations faster. The other way is to augment the available training data. Few-shot learning models, which aim for faster generalizations through architectural alterations, usually transfer knowledge gained from large-scale source tasks (using huge datasets) to small-scale target tasks with limited training data availability (Hu et al. 2022). Since body maps© only contain sparse information (see Figure 1), transferring gained knowledge from large-scale source tasks utilizing huge amounts of dense real-world images is difficult. Therefore, augmenting the available training data appears more feasible.

Here, the first effective data augmentation techniques were simple transformations such as horizontal flipping, color space augmentations, or random cropping (Krizhevsky et al. 2012). However, as body maps© are standardized and digitally created, applying such transformations would not add any realistic or expectable variation to the training data. More advanced deep learning based approaches such as using GANs or Diffusion Models to generate synthetic data have been proven to be effective (Frid-Adar et al. 2018, Trabucco et al. 2023). However, training GANs or Diffusion Models requires huge amounts of training data, which in our case is not available for filled-out body maps©. Therefore, we finally decided to augment our limited available training data by algorithmically generating images that represent filled-out body maps©.

## CONCLUSION

In our work, we investigated the effectiveness of synthetic data in the domain of canine musculoskeletal diagnosis. Here, the use of body maps© to visually document a canine's condition has the potential to advance canine musculoskeletal assessment. Our investigation of mitigating data scarcity through the use of synthetic data led to a notable improvement of around 10% in diagnosis accuracy on a smaller evaluation dataset, showcasing the potential of pre-training an AI model with generated synthetic images. While the positive impact on the diagnosis accuracy by using synthetic data diminished for a larger dataset, our findings still show the potential of synthetic data, especially when training data is very limited. This research not only contributes to advancing AI-based diagnostic support systems for canine health but also offers insight into broader applications facing data scarcity.

## REFERENCES

Appleby, R. B., & Basran, P. S. (2022). Artificial intelligence in veterinary medicine. Journal of the American Veterinary Medical Association, 260(8), p. 819–824.

Burg, M. F., Wenzel, F., Zietlow, D., Horn, M., Makansi, O., Locatello, F., & Russell, C. (2023). A data augmentation perspective on diffusion models and retrieval. Unpublished. Retrieved January 19, 2024, from https://arxiv.org/pdf/2304.10253.pdf.

Cornwall, A. (1992). Body mapping in health RRA/PRA. In RRA Notes, 16(July), p. 69-76.

Deng, J., Dong, W., Socher, R., Li, L. -J., Li, K., & Li Fei-Fei. (2009). ImageNet: A large-scale hierarchical image database. In 2009 IEEE Conference on Computer Vision and Pattern Recognition, p. 248-255.

European Physical and Rehabilitation Medicine Bodies Alliance. White Book on Physical and Rehabilitation Medicine (PRM) in Europe. In European Journal of Physical and Rehabilitation Medicine (2018), 54(2), p. 125-155 and 230-260.

Flutter - Build apps for any screen (2024). Retrieved January 19, 2024, from https://flutter.dev/.

Fortney, W. D. (2012). Implementing a successful senior/geriatric health care program for veterinarians, veterinary technicians, and office managers. In The Veterinary clinics of North America: Small animal practice, 42(4), p. 823-834.

Frid-Adar, M., Diamant, I., Klang, E., Amitai, M., Goldberger, J., & Greenspan, H. (2018). GAN-based synthetic medical image augmentation for increased CNN performance in liver lesion classification. Neurocomputing, 321, p. 321–331.

Galloway, S. S. (2011). How to document a dental examination and procedure using a dental chart. In American Association of Equine Practitioners (pp. 35-49).

Goodfellow, I., Pouget-Abadie, J., Mirza, M., Xu, B., Warde-Farley, D., Ozair, S., Courville, A., & Bengio, Y. (2014). Generative Adversarial Nets. In Advances in Neural Information Processing Systems.

Gysin, M., Tschuor, F., Schweizer, D. E., Nett, C., & Geissbühler, U. (2019). Stand der Digitalisierung in Schweizer Tierarztpraxen. In Schweizer Archiv für Tierheilkunde, 161(5), p. 307–317.

Ho, J., Jain, A., & Abbeel, P. (2020). Denoising Diffusion Probabilistic Models. In Advances in Neural Information Processing Systems, p. 6840–6851.

Krizhevsky, A., Sutskever, I., & Hinton, G. (2012). ImageNet Classification with Deep Convolutional Neural Networks. In Advances in Neural Information Processing Systems.

Lustgarten, J. L., Zehnder, A., Shipman, W., Gancher, E., & Webb, T. L. (2020). Veterinary informatics: forging the future between veterinary medicine, human medicine, and One Health initiatives-a joint paper by the Association for Veterinary Informatics (AVI) and the CTSA One Health Alliance (COHA). In JAMIA open, 3(2), p. 306–317.

Ratsch, B. E., Levine, D., Wakshlag J. J. (2012). Clinical Guide to Obesity and Nonherbal Nutraceuticals in Canine Orthopedic Conditions. In The Veterinary clinics of North America: Small animal practice, 52(4), p. 939-958.

Sheng, Z., Bollig, E., Granick, J., Zhang, R., & Beaudoin, A. (2022). Canine Parvovirus Diagnosis Classification Utilizing Veterinary Free-Text Notes. In 2022 IEEE 10th International Conference on Healthcare Informatics, p. 614–615.

Tan, M., & Le, Q. (2021). EfficientNetV2: Smaller Models and Faster Training. In Proceedings of the 38th International Conference on Machine Learning, p. 10096–10106.

Tarr, J., & Thomas, H. (2011). Mapping embodiment: Methodologies for representing pain and injury. In Qualitative research, 11(2), p. 141-157.

Trabucco, B., Doherty, K., Gurinas, M., & Salakhutdinov, R. (2023). Effective Data Augmentation With Diffusion Models. In ICLR 2023 Workshop on Mathematical and Empirical Understanding of Foundation Models.